\documentclass[11pt,a4paper]{article}
\usepackage[draft]{hyperref}
\usepackage{times,latexsym}
\usepackage{url}
\usepackage[T1]{fontenc}
\usepackage{listings}
\usepackage{tablefootnote,footnotehyper}
\usepackage{algorithm2e}
\usepackage{color, colortbl}
\usepackage[acceptedWithA]{tacl2018v2}
\usepackage{url}
\usepackage{amssymb}
\usepackage{booktabs} 
\usepackage{enumitem}
\usepackage{multirow}
\usepackage{graphicx}
\usepackage{subfigure}
\usepackage{amsmath}
\usepackage{floatrow}

\usepackage{siunitx}

\usepackage{xspace,mfirstuc,tabulary}
\newcommand{\dateOfLastUpdate}{Sept. 20, 2018}
\newcommand{\styleFileVersion}{tacl2018v2}

\newcommand{\ex}[1]{{\sf #1}}

\newif\iftaclinstructions
\taclinstructionsfalse 
\iftaclinstructions
\renewcommand{\confidential}{}
\renewcommand{\anonsubtext}{(No author info supplied here, for consistency with
TACL-submission anonymization requirements)}
\newcommand{\instr}
\fi

\iftaclpubformat 
\newcommand{\taclpaper}{final version\xspace}

\newcommand{\Taclpaper}{Final version\xspace}
\newcommand{\Taclpapers}{Final versions\xspace}

\else
\newcommand{\taclpaper}{submission\xspace}

\newcommand{\Taclpaper}{Submission\xspace}
\newcommand{\Taclpapers}{{\Taclpaper}s\xspace}

\fi

\title{The Source-Target Domain Mismatch Problem in Machine Translation}

\author{
Jiajun Shen$^{\diamond\dagger}$ \enskip Peng-Jen Chen$^{\diamond\dagger}$ \enskip Matt Le$^{\dagger}$ \enskip Junxian He$^{\bullet}$\Thanks{Work done while internship at the Facebook AI Research lab.} \enskip Jiatao Gu$^{\dagger}$ \\ 
\enskip {\bf Myle Ott}$^{\dagger}$ \enskip {\bf Michael Auli}$^{\dagger}$ \enskip {\bf Marc'Aurelio Ranzato}$^{\dagger}$\\
$^\dagger$Facebook AI Research \\
$^\bullet$Carnergie Mellon University \\
{\sf \{jiajunshen,pipibjc,mattle,jgu,myleott,michaelauli,ranzato\}@fb.com} \enskip {\sf junxianh@cs.cmu.edu} \\
}

\date{}

\begin{document}
\maketitle

\begin{abstract}
While we live in an increasingly interconnected world, different places still exhibit strikingly different cultures and many events we experience in our every day life pertain only to the specific place we live in. As a result, people often talk about different things in different parts of the world. 
In this work we study the effect of local context in machine translation and postulate that particularly in low resource settings this causes the domains of the source and target language to greatly mismatch, as the two languages are often spoken in further apart regions of the world with more distinctive cultural traits and unrelated local events. 
We first formalize the concept of source-target domain mismatch, propose a metric to quantify it, and provide empirical evidence 
corroborating our intuition that organic text produced by people speaking very different languages exhibits the most dramatic differences.
We conclude with an empirical study of how source-target domain mismatch affects training of machine translation systems for 
low resource language pairs. In particular, we find that it severely affects back-translation, 
but the degradation can be alleviated by combining back-translation with self-training and by increasing the relative amount of target side
monolingual data.
\end{abstract}

\renewcommand{\thefootnote}{$^{\diamond}$}
\footnotetext[1]{Equal contribution.}
\renewcommand\thefootnote{\arabic{footnote}}

\section{Introduction}
The use of language greatly varies with the geographic location~\citep{firth35,johnstone}. 
Even within places where people speak the same language~\citep{britain}, there is a lot of lexical variability 
due to change of style and topic distribution, particularly when considering content posted on social media, 
blogs and news outlets. 
For instance, while a primary topic of discussion between British sport fans is cricket, American sport fans are more likely to discuss other sports such as baseball~\citep{leech}. 

The effect of local context in the use of language is even more extreme when considering regions where different languages are spoken. Despite the increasingly interconnected world we live in, people in different places tend to talk about different things. There are several reasons for this, from cultural differences due to geographic separation and history, to the local nature of many events we experience in our every day life; e.g., the traffic congestion in Taipei is not affected by a heavy snowfall in New York City. 

This phenomenon has not only interesting socio-linguistic aspects but it has also strong implications in machine translation~\citep{bernardini04}. 
In particular, machine translation of low-resource language pairs aims at automatically translating content in two languages that are often spoken in very distant geographic locations by people with rather different cultures. In machine learning terms
and at a very high level of abstraction, this is akin to the problem of {\em aligning} two very high dimensional and sparsely populated point clouds. The learning problem is difficult because not only very few correspondences are provided to the learner,
 but also because the distributions of points is rather different.

As of today, most machine translation research has been based on the often implicit assumption that 
content in the two languages is {\em comparable}. Sentences comprising the parallel dataset used for training are assumed to cover
the same topic distribution, regardless of the originating language. 
Similarly, monolingual corpora are assumed to be comparable, i.e. to cover the same distribution of topics albeit in 
two different languages. 

The major contribution of this work is to raise awareness in the machine translation community that 
 this assumption does not hold for the vast majority of language pairs, which are distant and low-resource, and for the vast
majority of the content produced every day on the Internet by means of blogs, social platforms and news outlets.

In \textsection\ref{sec:stdm}, we first propose a formal definition of source-target domain mismatch (STDM). 
This abstraction precisely characterizes the
problem and exposes the assumptions needed to formulate a practical definition of a metric, which we dub \emph{STDM score} and
describe in \textsection\ref{sec:metric}.  
The STDM score quantifies the degree of domain mismatch between a set of parallel sentences 
originating in the source and target language. Empirically, this score indicates an overall larger mismatch for
data originating in more distant language and for more organic content, like the one derived from social media data;
see \textsection\ref{sec:stdm_score_ds} for details. This suggests that applying methods proven to work well on most popular WMT 
benchmarks may generalize poorly to less constrained settings and low resource languages.

Therefore, we conclude by analyzing the consequences of STDM on low resource machine translation in \textsection\ref{sec:baselines}.
We surmise that STDM may negatively impact the
effectiveness of back-translation~\citep{sennrich2015improving}, which is {\em de facto} the best known approach 
to leverage monolingual data in low resource settings. 
In particular, even if the backward model was perfect, back-translation may be less effective when there is considerable STDM, 
since the back-translated data is out-of-domain relative to the source domain from which we aim to translate.

To validate this conjecture, in \textsection\ref{sec:results} 
we work with a synthetic benchmark that enables us to precisely control the amount of STDM.
We then assess the effectiveness of back-translation as a function of the amount of STDM, as well as other factors such as 
the amount of data available.
We find that back-translation is sensitive to STDM, but this can be compensated by adding more target-side monolingual data and by combining back-translation 
with self-training~\citep{yarowski}. In \textsection\ref{sec:lrmt} we confirm our findings on two actual low resource language pairs, 
Nepali-English and English-Myanmar.

Our conclusion is that STDM is an intrinsic property of the translation task, particularly for distant languages and 
uncurated content. 
In these conditions,  STDM can affect generalization of MT systems, but the degradation
depends on several factors, such as the amount of data originating in each language and the particular language pair.


\section{Related Work} \label{sec:related}
The observation that topic distributions and various kinds of lexical variabilities depend on the local context has been 
known and studied for a long time. For instance, \citet{firth35} says ``{\em Most of the give-and-take of conversation in our everyday life is stereotyped and very narrowly conditioned by our particular type of culture}''.
In her seminal work, \citet{johnstone} analyzed the role of place in language, focusing on lexical variations within 
the same language, a subject further explored by~\citet{britain}. 
Some of these works were the basis for later studies that introduced computational models for 
how language changes with geographic location~\citep{mei06, eisenstein10}. 

Moving to cross-lingual analyses, there has been work at the intersection of linguistics and cognitive science~\citep{Pederson}
 showing how certain linguistic codings vary across languages, and how these affect how people form mental concepts.
In the field of topic modeling, there has been a new sub-field emerging over the past 10 years focusing on modeling multi-lingual corpora~\citep{mimno09,graber09,gutierrez16}.
However, only recently had researchers dropped assumptions on the use of parallel and comparable corpora~\citep{hao18, yang19}.
While some works do investigate issues related to STDM~\citep{gutierrez16}, like how named entities receive a different distribution over words in different languages~\citep{lin18}, none of these works have analyzed how the overall topic distribution of data originating in the source and target language differ.

In machine translation, researchers have often made an explicit assumption on the use of {\em comparable} corpora~\citep{fung98,marcu04,irvine13}, i.e. corpora in the two languages that roughly cover the same set of topics. 
Unfortunately, monolingual corpora are seldom comparable in practice. \citet{leech} analyzes two comparable corpora, one in American English and the other in British English, and demonstrate differences that reflect the cultures of origin. Similarly,
\citet{bernardini04} observes that parallel datasets built for machine translation exhibit strong biases in the 
selection of the original documents, making the text collection not quite comparable. 

The non-comparable nature of machine translation datasets is even more striking when considering low resource language pairs, 
for which differences in local context and cultures are more pronounced.
Recent studies~\citep{sogaard18, neubig18emnlp} have warned that removing the assumption on comparable corpora strongly 
deteriorates performance of  lexicon induction techniques which are at the foundation of machine translation.
 
To the best of our knowledge, no prior work has so far made explicit the intrinsic mismatch between source and target 
domain in machine translation, both when considering the portion of the parallel dataset originating in the source and 
target language, and when considering the source and target monolingual corpora. We believe that this is an important
characteristic of machine translation tasks, particularly when the content is derived from blogs, social media platforms,
and news outlets. In fact, any attempt at making corpora comparable would change the nature of the original task, as we are
usually interested in translating content originating in the source language.

Back-translation~\citep{sennrich2015improving} has been the workhorse of modern neural MT, 
enabling very effective use of target side monolingual
data. Back-translation is beneficial because it helps regularizing the model and adapting to new domains~\citep{burlot18}. 
However, the typical
setting of current MT benchmarks as popularized by recent WMT competitions~\cite{bojar2019wmt} is a mismatch 
between {\em training and test} sets, as opposed to a mismatch between {\em source and target} domains as in this work. 
In this setting, vast amounts of target monolingual data in the domain of the test set can be leveraged 
very effectively by back-translation. Unfortunately, back-translation
is much less effective when dealing with STDM, as we will show in \textsection\ref{sec:mt_controlled}. \citet{zheng-etal-2019-robust}
tackles this problem by adding tags to examples~\citep{caswell-etal-2019-tagged} to let the model know whether the data
originates from the source or target domain. We employ this technique also in our experiments.

There has been some work attempting to make better use of source side monolingual data, 
as this is in-domain with the text we would like to translate at test time. 
\citet{ueffing} proposed to improve a statistical MT system using {\em self-training}~\citep{yarowski}, 
a direction later pursued by~\citet{zhang16} for neural MT. 
In our work, we consider the iterative variant proposed by~\citet{st_he19}, whereby all model parameters are subject to training and
 noise is added to the input. \citet{chinea} showed that self-training can be used to adapt to a different domain after
selecting from a source monolingal dataset sentences that are similar to the test domain. 
\citet{guanlinli} compares back-translation and self-training with respect to input sensitivity and prediction
margin. None of this works however analyze how these methods fair when there is source-target domain mismatch which is the focus of this work.
In this work, we also report improvements when combining self-training 
with back-translation. This is consistent with earlier findings by~\citet{DBLP:journals/corr/ParkSY17}, who however 
combined forward and back-traslated data to alleviate biases in the corresponding MT systems as opposed to compensate for domain effects.

\citet{Kilgarriff98} proposed a controlled setting to study metrics to assess similarity between corpora in the same language by 
defining a mixture between two known corpora. 
In \textsection\ref{sec:contr}, we will use the same method but we apply it to corpora in two languages as required for machine translation.
Finally, \citet{Fothergill16} also defines a metric in the topic space, albeit for corpora in the same language. In our case, working in the 
topic space makes our measures more robust to translationese effects~\citep{zhang2019translationese}, which could otherwise be a greater confounding factor in the assessment of STDM (\textsection\ref{sec:translationese}).

\section{The STDM Problem} \label{sec:stdm}
\begin{figure}[!t]
\begin{center}
\includegraphics[width=1\linewidth]{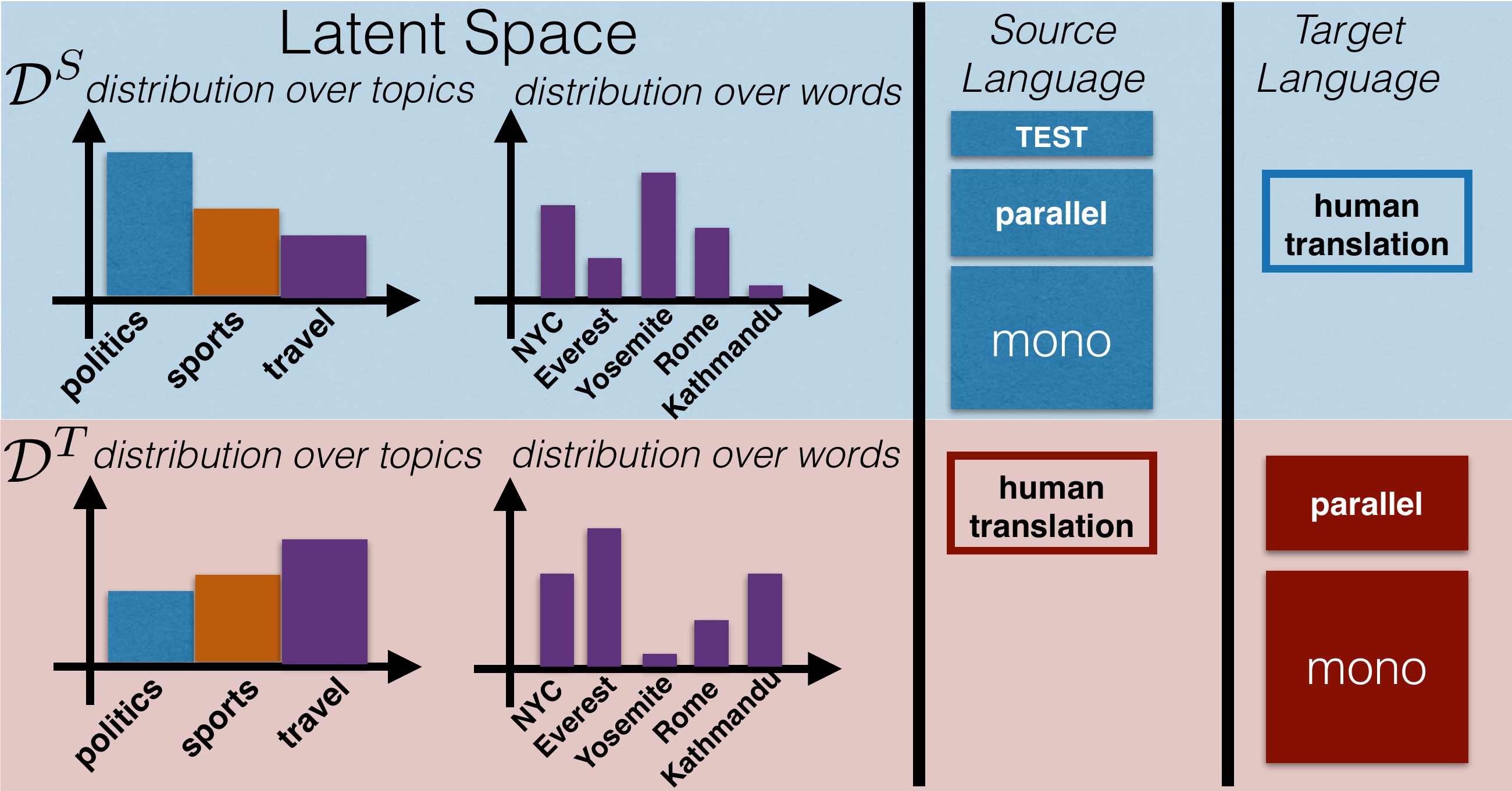}
\end{center}
\caption{\small
Toy illustration of STDM in MT. There are two domains, the source domain $\mathcal{D}^S$ (top) and the target domain $\mathcal{D}^T$ (bottom).
We postulate that in a latent concept space these two domains differ because the topic distributions are different (e.g., 
in the source domain politics is more popular than travel) and because for the same topic the word distributions are different (e.g., 
the word ``Everest'' is more common than ``Yosemite''  in the travel topic of the target domain). 
On the right hand side, we show how STDM manifests in machine translation datasets.
All data originating in the source language belongs to the source domain, this includes a portion of the parallel dataset, the source side
monolingual dataset and the test set we eventually would like to translate. Empty boxes represent human translated data in the 
parallel training dataset.}
\label{fig:outline2}
\vspace{-.2cm}
\end{figure}
In this section we formalize the definition of Source-Target Domain Mismatch (STDM); this is an intrinsic property of 
the data which is independent of the particular machine translation system under consideration.
We assume there exists a latent concept space shared across all languages. The process to generate a sentence follows 
the standard data generation process used in topic modeling, whereby we first sample a distribution over topics, $\pi_i \sim \Pi$
where $i$ is an index over topics, and then a distribution  over words for each topic, $w_{ij} \sim \pi_i$, where $j$ indexes the words
in the dictionary. Next, we assume there are two distinct domains, 
the source domain $\mathcal{D}^S$ and the target domain $\mathcal{D}^T$. These two domains differ in both the distribution
over topics $\Pi$, and the distribution over words given a certain topic $\pi_i$, as depicted in Fig.~\ref{fig:outline2}.
For the sake of conciseness, we will refer to $z^s$ and $z^t$ as sentences in the concept space generated from domain $\mathcal{D}^S$
and $\mathcal{D}^T$, respectively.

Let's imagine now that we have generated two sets of sentences in each domain. What we observe in practice is their realization in 
each language, $\mbox{src}(z^s)$ and $\mbox{tgt}(z^t)$, where $\mbox{src}$ and $\mbox{tgt}$ map sentences from the concept space to
the source and target language, respectively. Finally, let's denote with $h_{s \rightarrow t}$ and $h_{t \rightarrow s}$ the functions
representing human translations of source sentences in the target language and vice versa.

In the simplest setting, a machine translation dataset is composed of parallel and monolingual datasets.
Using the notation introduced above, the parallel dataset is denoted by $\mathcal{P} = \{ (\mbox{src}(z^s), 
h_{s \rightarrow t}(\mbox{src}(z^s)) \}_{z^s \sim \mathcal{D}^S} \cup  \{ (h_{t \rightarrow s}(\mbox{tgt}(z^t)), 
\mbox{tgt}(z^t)  \}_{z^t \sim \mathcal{D}^T}$. The first set originates in the source language and belongs to the source domain, while
the second set originates in the target language and belongs to the target domain. We then have a source side monolingual dataset, 
$\mathcal{M}^S = \{ \mbox{src}(z^s) \}_{z^s \sim \mathcal{D}^S}$, and a target side monolingual dataset, 
$\mathcal{M}^T = \{ \mbox{tgt}(z^t) \}_{z^t \sim \mathcal{D}^T}$, 
belonging to the source and target domains, respectively.
The {\em test set} which we would like
to eventually translate contains sentences in the source language, all belonging to the {\em source} domain. The existence of distinct source
and target domains and datasets derived from these two domains as described above define the STDM problem.

In most domain adaptation studies for machine translation, it is assumed that $\mathcal{D}^S = \mathcal{D}^T$ but the test domain differs 
from the training domain. Here instead, the test domain is in the same domain as the portion of the training set originating 
in the source language. We would like to a) understand the effects of such mismatch and b) understand how to best  
leverage the out-of-domain data originating from the target language (target monolingual dataset and portion of the parallel dataset originating in the target language).

We conclude with a disclaimer for the critical reader. 
In reality, there may not exist a shared concept space across all languages, since some concepts may be unique to a language. Moreover, the granularity of how topics are defined is arbitrary. 
Finally, in practice there may be not two but multiple domains and multiple languages.
Despite these limitations and assumptions, 
we will show in the following sections that this simple framework has reasonable empirical support and that it
can help us define a useful metric. We will analyze the implications for learning machine translation systems in 
\textsection\ref{sec:baselines}.

\subsection{Empirical Evidence}
\begin{figure}[t]
\begin{center}
\includegraphics[scale=0.5]{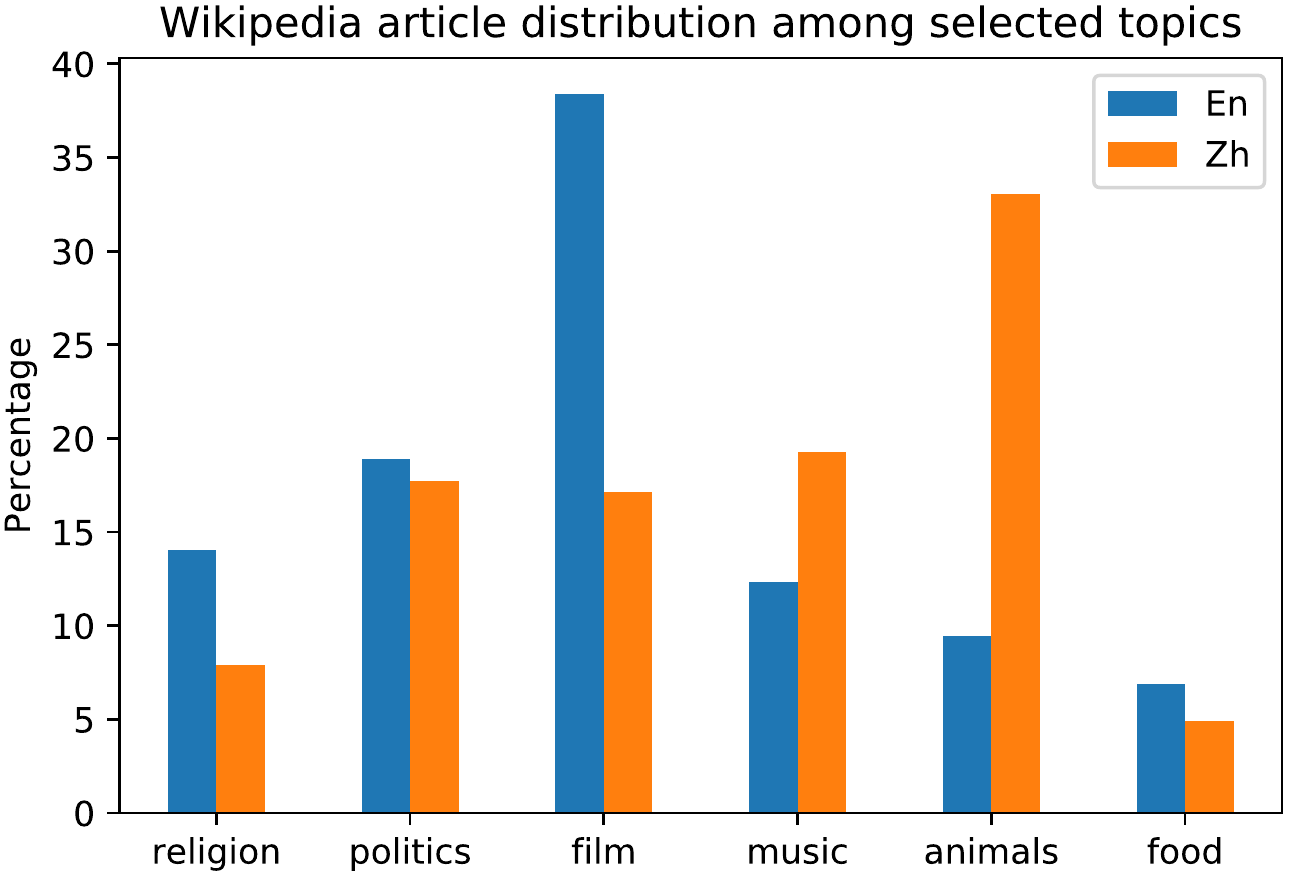}
\end{center}
\caption{\small Topic distribution of Wikipedia pages written in English and Chinese.}
\label{fig:topic_en_zh_wiki}
\vspace{-.3cm}
\end{figure}
In this section we first provide anecdotal evidence that documents originating in different languages possess different distributions over
topics. We train two topic classifiers, one for Chinese and the other for English, using the Wikipedia annotated data  
from~\citet{yuan19}. We apply this classifier to 20,000 documents randomly sampled from English and Chinese Wikipedia.
Fig.~\ref{fig:topic_en_zh_wiki} shows that according to this classifier, English Wikipedia has more pages about entertainment and 
religion than Chinese Wikipedia, for instance.

Second, we provide empirical support for the claim that corpora originating in different places may have different word distributions for
the same set of topics. Towards this end, we summarize \citet{leech}'s seminal study who analyzed the Brown corpus and the LOB corpus of 
British and American English text, respectively. These are examples of corpora comprising text extracted from the same proportion of 
text categories and using essentially the same sampling procedure for their construction. Yet the authors find a different usage of 
vocabulary, particularly for gender related words. The authors conclude that ``\emph{... we may propose a picture of US culture in 1961 -- masculine
to the point of machismo, militaristic, ... -- contrasting with one of British culture as more given to temporizing and talking... 
 and to family and emotional life...}''. All together, empirical evidence suggests that
STDM can be attributed to both differences in the topic distribution as well as word distributions for the same topic.

\section{Metric: The STDM Score} \label{sec:metric}
Given the framework introduced in \textsection\ref{sec:stdm}, in this section we are going to discuss a practical way to measure STDM.
Ideally, we would like to measure a distance between two sample distributions, $z^s \sim \mathcal{D}^S$ and $z^t \sim \mathcal{D}^T$.
Unfortunately, we have no access to such latent space. What we observe are realizations in the source and target language.
However, it is also an open research question~\citep{hao18,yang19} how to compare 
the distribution of $\{\mbox{src}(z^s)\}$ against $\{\mbox{tgt}(z^t)\}$, 
since these are two possibly incomparable corpora in different languages.

In this work, we therefore leverage the existence of a parallel corpus and compare the distribution of 
$\mathcal{A}^T = \{\mbox{tgt}(z^t)\}_{z^t \sim \mathcal{D}^T}$ with
$\mathcal{A}^S = \{h_{s \rightarrow t}(\mbox{src}(z^s))\}_{z^s \sim \mathcal{D}^S}$. The underlying assumption is that the effect of
translationese~\citep{baker1993translationese,zhang2019translationese,toury2012translationstudies} 
is negligible compared to the actual STDM, and therefore, we can ignore changes to the distribution brought 
by the mapping $h_{s \rightarrow t}$. We will validate this assumption in \textsection\ref{sec:translationese}.

Next, we assume that what contributes the most to STDM are changes between the topic distributions of source and target domains.
Under this additional assumption, we define the score as a measure of the topic discrepancy between $\mathcal{A}^S$ and $\mathcal{A}^T$.
Let $\mathcal{A} = \mathcal{A}^S \cup \mathcal{A}^T$ be the concatenation of the corpus originating in the source and target language.
We first extract topics using LSA (but any other method could be considered). Let $A \in \mathbb{R}^{(n^S+n^T) \times k}$ be the TF-IDF matrix derived from $\mathcal{A}$ where the first
$n^S$ rows are representations taken from $\mathcal{A}^S$, the bottom $n^T$ rows are representations of $\mathcal{A}^T$, and $k$ is the 
number of words in the dictionary. The SVD decomposition of $A$ yields: $A = U S V = (U \sqrt{(S)}) (\sqrt{(S)} V) = \bar{U} \bar{V}$. 
Matrix $\bar{U}$ collects topic representations of the original documents; let's denote by $\bar{U}^S$ the first $n^S$ rows corresponding to
$\mathcal{A}^S$ and $\bar{U}^T$ the remaining $n^T$ rows corresponding to $\mathcal{A}^T$. 
Let $C = \bar{U} \bar{U}' =  \big[\begin{smallmatrix} C^{SS} & C^{ST} \\ C^{ST}{'} & C^{TT}
\end{smallmatrix}\big]$, 
where $C^{SS} = \bar{U}^S \bar{U}^S{'}$, $C^{ST} = \bar{U}^S \bar{U}^T{'}$ and $C^{TT} = \bar{U}^T \bar{U}^T{'}$.
The STDM score is defined as:
\begin{equation}
\mbox{score}  =  \frac{s^{\mbox{\tiny ST}} + s^{\mbox{\tiny TS}}}{s^{\mbox{\tiny SS}} + s^{\mbox{\tiny TT}}}, \mbox{\small with }
s^{\mbox{\tiny AB}}  =  \frac{1}{n^A n^B} \sum_{i=1}^{n^A} \sum_{j=1}^{n^B} C_{i,j}^{AB} \label{eq:score} \\\nonumber
\end{equation} 
where $s^{\mbox{\tiny AB}}$ measures the average similarity between documents of set A 
to documents of set B. The score measures the  cross-corpus similarity normalized by the within corpus similarity.
In the extreme setting where $\mathcal{D}^S$ and $\mathcal{D}^T$ are fully disjoint, then we would have that the off-diagonal block $C^{ST}$
is going to be a zero matrix and therefore the score is equal to 0. When the two domains perfectly match instead, $s^{\mbox{\tiny SS}}=
s^{\mbox{\tiny TT}}=s^{\mbox{\tiny ST}}=s^{\mbox{\tiny TS}}$, and therefore, the score is equal to 1. In practice, we expect a score in the range $[0, 1]$.

\subsection{A Controlled Setting}  \label{sec:contr}
Similarly to~\citet{Kilgarriff98}, we introduce a synthetic benchmark to {\em finely control} the domain of the 
target originating data, and therefore the amount of STDM. The objective is to assess whether the STDM score defined in Eq.~\ref{eq:score}
captures well the expected amount of mismatch.

The key idea of this controlled setting is to use a convex combination of data from two sufficiently different domains as target originating data, which comprises the target side monolingual data and half of the parallel training data.

In this work we use EuroParl~\citep{europarl} as our source originating data, while our target originating data contains a mix of data 
from EuroParl and OpenSubtitles~\citep{opensub}.
Specifically, we consider a French to English translation task with a parallel dataset composed of 10,000 sentences 
from EuroParl (which is assumed to originate in
French) and 10,000 sentences from the target domain (which is assumed to originate in English).

Let $\alpha \in [0, 1]$, the domain of the target originating data is set to: $\alpha$ EuroParl + $(1-\alpha)$ OpenSubtitles. 
For instance, when $\alpha=0$ then the target domain (OpenSubtitles) is totally out-of-domain with respect to the source domain (EuroParl).
When $\alpha=1$ instead, the target domain matches perfectly the source domain.
For intermediate values of $\alpha$, the match is only partial.
Notice that even when $\alpha=0$, we assume that the parallel dataset is comprised of two halves, one originating from the EuroParl domain
(the ``French originating'' data) and one from OpenSubtitles (the ``English originating'' data).

Next, we evaluate the STDM score as a function of $\alpha$. As we can see from Table~\ref{tab:stdm_score_controlled} and as expected, 
the STDM score increases fairly linearly as we increase the value of $\alpha$. 
\begin{table}[t]
\centering
\small
\begin{tabular}{l|ccccc}
$\alpha$ & 0 & 0.25 & 0.5 & 0.75 & 1.0 \\
\hline
STDM score & 0.29 & 0.55 & 0.78 & 0.93 & 0.99 \\
\end{tabular}
\caption{\small STDM score as a function of the parameter $\alpha$ controlling the STDM in the synthetic setting.}
\label{tab:stdm_score_controlled}
\end{table}

\subsection{Empirical Evaluation of STDM on Various Datasets} \label{sec:stdm_score_ds}
\begin{table}[t]
\centering
\scriptsize
\begin{tabular}{l||c|c|c|c|c|c}
      & De-En & Fi-En & Ru-En & Ne-En & Zh-En & Ja-En \\
\hline
WMT   & 0.79  & 0.79  & 0.76  & -     & 0.65  & - \\
MTNT &  -    &   -   &  -    &  -    & -     & 0.69 \\
SMD  & 0.81  & 0.71  & 0.71  & 0.64  & 0.71  & 0.61 \\
\end{tabular}
\caption{\small STDM score on several language pairs using parallel data from WMT, MTNT and from a social media platform (SMD) test sets.}
\label{tab:stdm_score_real_ds}
\end{table}

We now evaluate the STDM score on real data.
We consider six language pairs, German-English, Finnish-English, Russian-English, Nepali-English, Chinese-English and Japanese-English.
We analyze datasets from WMT, MTNT~\citep{michel2018mtnt} and from a social media platform (SMD). 
For each language, we sample 5000 sentences from WMT newstest sets and MTNT dataset, and 20000 sentences from SMD. We then merge all these datasets and their English translations to compute a common set of topics, making STDM scores comparable across language pairs and datasets.

The results in Table~\ref{tab:stdm_score_real_ds} are striking. 
First, WMT datasets, except for Chinese, show relatively mild signs of STDM and negligible difference across language pairs, suggesting that the data curation process of WMT datasets have made source and target originating corpora rather comparable. 
The distribution of WMT Chinese originating data instead is rather different because it contains much more local news, while the other 
languages are mostly about international news which are largely language independent. 
Interestingly, En-De data derived from social media data has even milder STDM, Fi-En and Ru-En have more substantial STDM. 
Instead, \emph{MTNT and SMD exhibit strong signs of STDM for distant languages} like Nepali, Chinese and Japanese. 
This agrees well with our intuition that STDM is more severe for more distant languages associated to more diverse cultures.

\subsubsection{The Effect of Translationese} \label{sec:translationese}
In \textsection\ref{sec:stdm} we have made the assumption that the effect of translationese is negligible when estimating STDM.
However, there are previous studies showing clear artifacts in (human) 
translations~\citep{baker1993translationese,zhang2019translationese,toury2012translationstudies}. In this section we aim at assessing
whether our STDM score is affected by translationese.

We consider the WMT'17 De-En dataset from~\citet{ott_icml18} which contains double translations of source and target originating sentences.
From this, we construct paired inputs and labels, 
$\{ ( h_{s \rightarrow t}( h_{t \rightarrow s}(\mbox{tgt}(z^t))),1)\} \cup \{ (\mbox{tgt}(z^t), 0) \} $, and train two classifiers to predict whether or not the input is translationese.
The first classifier takes as input a TF-IDF representation
$w$ of the sentence, while the second classifier takes only the corresponding topic distribution: $\bar{V} w$.
On this binary task a linear classifier achieves 58\% accuracy on the test set with TF-IDF input representations, and only 52\% when given just the topic distribution. If we apply the same binary classifier in the topic space to discriminate between sentences originating in the source and target domain ($\mbox{tgt}(z^t)$ VS. $h_{s \rightarrow t}( \mbox{src}(z^s) )$), the accuracy increases to 64\%.

We conclude that once we control for domain effect (by discriminating the same set of sentences in their original form versus their double
translationese form), the accuracy is much lower than previously reported~\citep{zhang2019translationese}, and working in the
topic space further removes translationese artifacts. Therefore, the STDM score computed in the topic space is unlikely affected by such
artifacts and captures the desired discrepancy between the source and the target domains.

\section{The Effect of STDM in Machine Translation} \label{sec:baselines}
In this section, we turn our attention to how STDM affects training of machine translation systems.
We consider state-of-the-art neural machine translation (NMT) systems based on 
the transformer architecture~\citep{vaswani2017transformer} with subword vocabularies learned 
via byte-pair encoding (BPE)~\cite{sennrich2015improving}. In order to adapt to the different domains, we employ domain 
tagging~\citep{zheng-etal-2019-robust} by adding a domain token to the input source sentence\footnote{In the controlled setting of 
\textsection\ref{sec:mt_controlled} we found that tagging a small but consistent improvement by almost 1 BLEU point.}.
We also use label smoothing~\citep{labelsmoothing} and dropout~\citep{dropout} to improve generalization, as we focus
on low resource language pairs where models tend to severely overfit. Finally, we explore
ways to leverage both target and source side monolingual data via back-translation and self-training which we review next.

We simplify our notation and denote with $x^s=\mbox{src}(z^s)$ and $y^t=\mbox{tgt}(z^t)$ the source and 
target originating sentences,  $y^s=h_{s \rightarrow t}(x^s)$ and $x^t=h_{t \rightarrow s}(y^t)$ the corresponding human translations,
and $\hat{y}^s$ and  $\hat{x}^t$ the corresponding machine translations. The superscript always specifies the domain. We assume access to
a parallel dataset $\mathcal{P} = \{(x^s, y^s)\} \cup \{(x^t, y^t)\}$, a source side monolingual dataset $\mathcal{M}^s = \{x^s\}$ and a 
target side monolingual dataset $\mathcal{M}^t = \{y^t\}$.

\subsection{Back-Translation (BT)} \label{sec:bt}
Back-translation (BT)~\citep{sennrich2015improving} is a very effective data augmentation technique that leverages $\mathcal{M}^t$. 
The algorithm proceeds in three steps. First, a reverse machine translation
system is trained from target to source using the provided parallel data: 
$\overleftarrow{\theta} = \arg \max_{\theta} \mathbb{E}_{(x,y)\sim \mathcal{P}} \log p(x|y;\theta)$.
Then, the reverse model is used to translate the target monolingual data: 
$\hat{x}^t \approx \arg \max_z p( z | y^t; \overleftarrow{\theta})$, for $y^t \sim \mathcal{M}^t$.
The maximization is typically approximated by beam search.
Finally, the forward model is trained over the concatenation of the original parallel and back-translated data:
$\overrightarrow{\theta} = \arg \max_{\theta} \mathbb{E}_{(x,y)\sim \mathcal{Q}} \log p(y|x;\theta)$ with 
$\mathcal{Q} = \mathcal{P} \cup \{ \hat{x}^t, y^t \}_{y^t \sim \mathcal{M}^t}$. In practice,
the parallel data is weighted more in the loss, with a weight selected via hyper-parameter search on the validation set.

BT generally improves fluency and generalization, but has potential weaknesses when there is STDM.
Even if the reverse model were to produce perfect translations, 
back-translated data belongs to the target domain, and it is therefore out-of-domain with
the data we wish to translate, i.e., source sentences belonging to the source domain. We will verify this conjecture 
empirically in \textsection\ref{sec:mt_controlled}.

\subsection{Self-Training (ST)} \label{sec:st}

\begin{algorithm}[t]
\SetAlgoLined
\scriptsize
\nl \KwData{Given a parallel dataset $\mathcal{P}$ and
a source monolingual dataset $\mathcal{M}^s$ with $N^s$ examples\;}
\nl {\bf Noise:} Let $n(x)$ be a function that adds noise to the input by dropping, swapping and blanking words\;
\nl {\bf Hyper-params:} Let $k$ be the number of iterations and
 $A_1 < \dots < A_k \leq N_S$ be the number of samples to add at each iteration\;
\nl Train a forward model: $\overrightarrow{\theta} = \arg \max_{\theta}  \mathbb{E}_{(x,y)\sim \mathcal{P}} \log p(y|x;\theta)$\;
\nl \For{$t$ \textbf{in} $[1 \dots k]$}{
  \nl forward-translate data: $(\hat{y}^s, v) \approx \arg \max_z p( z | x^s; \overrightarrow{\theta})$, 
for $x^s \in \mathcal{M}^s$, where $v$ is the model score\;
  \nl Let $\bar{\mathcal{M}}^s \subset \mathcal{M}^s$ containing the top-$A_t$ highest scoring examples according to $v$\;
  \nl re-train forward model:
$\overrightarrow{\theta} = \arg \max_{\theta} \mathbb{E}_{(x,y)\sim \mathcal{Q}} \log p(y|x;\theta)$ with
$\mathcal{Q} = \mathcal{P} \cup \{n(x^s), \hat{y}^s\}_{x^s \sim \bar{\mathcal{M}}^s}$.}
\caption{Self-Training algorithm.
\label{algo:ST}}
\end{algorithm}

Self-Training (ST)~\citep{st_he19, yarowski}, shown in 
Alg.~\ref{algo:ST}, is another method for data augmentation that instead leverages  $\mathcal{M}^s$.
First, a baseline forward model is trained on the parallel data
(line 4). Second, this initial model is applied to the source monolingual data (line 6). Finally, the forward model
is re-trained from random initialization by augmenting the original parallel dataset with the forward-translated data.
As with BT, the parallel dataset receives more weight in the loss.

One benefit of this approach is that the synthetic parallel data added to the original parallel data 
is {\em in-domain}, unlike back-translated data.
However, the model may reinforce its own mistakes since synthetic targets are produced by the model itself.
Accordingly, we make the algorithm iterative and add only the examples for which the model was most confident (line 3, loop in line 5 and line 7). In our experiments we iterate three times.
We also inject noise to the input sentences, in the form of word swap and drop~\citep{lample_emnlp2018}, 
to further improve generalization (line 8). 

\subsection{Combining BT and ST} \label{sec:btst}
BT and ST are complementary to each other.
While BT benefits from correct targets, the synthetic data is out-of-domain when there is STDM.
Conversely, ST benefits from in-domain source sentences but synthetic targets may be inaccurate.
We therefore consider their combination as an additional baseline approach.

The combined learning algorithm proceeds in three steps. First, we train an initial forward and reverse model using the parallel 
dataset. Second, we back-translate target side monolingual data using the reverse model (see \textsection\ref{sec:bt}) and 
iteratively forward translate source side monolingual data using the forward model (see \textsection\ref{sec:st} and
Alg.~\ref{algo:ST}).
We then retrain the forward model from random initialization using the union of the original parallel dataset,
the synthetic back-translated data, and the synthetic forward translated data at the last iteration of the ST algorithm.

\section{Machine Translation Results} \label{sec:results}
In this section, we first study the effect of STDM on NMT using the controlled 
setting introduced in \textsection\ref{sec:contr} which enables us to assess the influence of 
various factors, such as the extent to which target originating data is out-of-domain, and the effect of monolingual data size.
We then report experiments on genuine low resource language pairs, namely Nepali-English and English-Myanmar.

We tune model hyperparameters (e.g., number of layers and hidden state size) and BPE size on the validation set.
Based on cross-validaiton, when training on datasets with less than 300k parallel sentences 
(including those from ST or BT), we use a 5-layer transformer with 8M parameters. 
The number of attention heads, embedding dimension and inner-layer dimension are 2, 256, 512, respectively.
When training on bigger datasets, we use a bigger transformer with 5 layers, 8 attention heads, 1024 embedding dimension, 2048 inner-layer dimension and a total of 110M parameters.
We report \textsc{SacreBLEU}~\citep{post-2018-call}.

\begin{figure}
\centering
\includegraphics[width=1\linewidth]{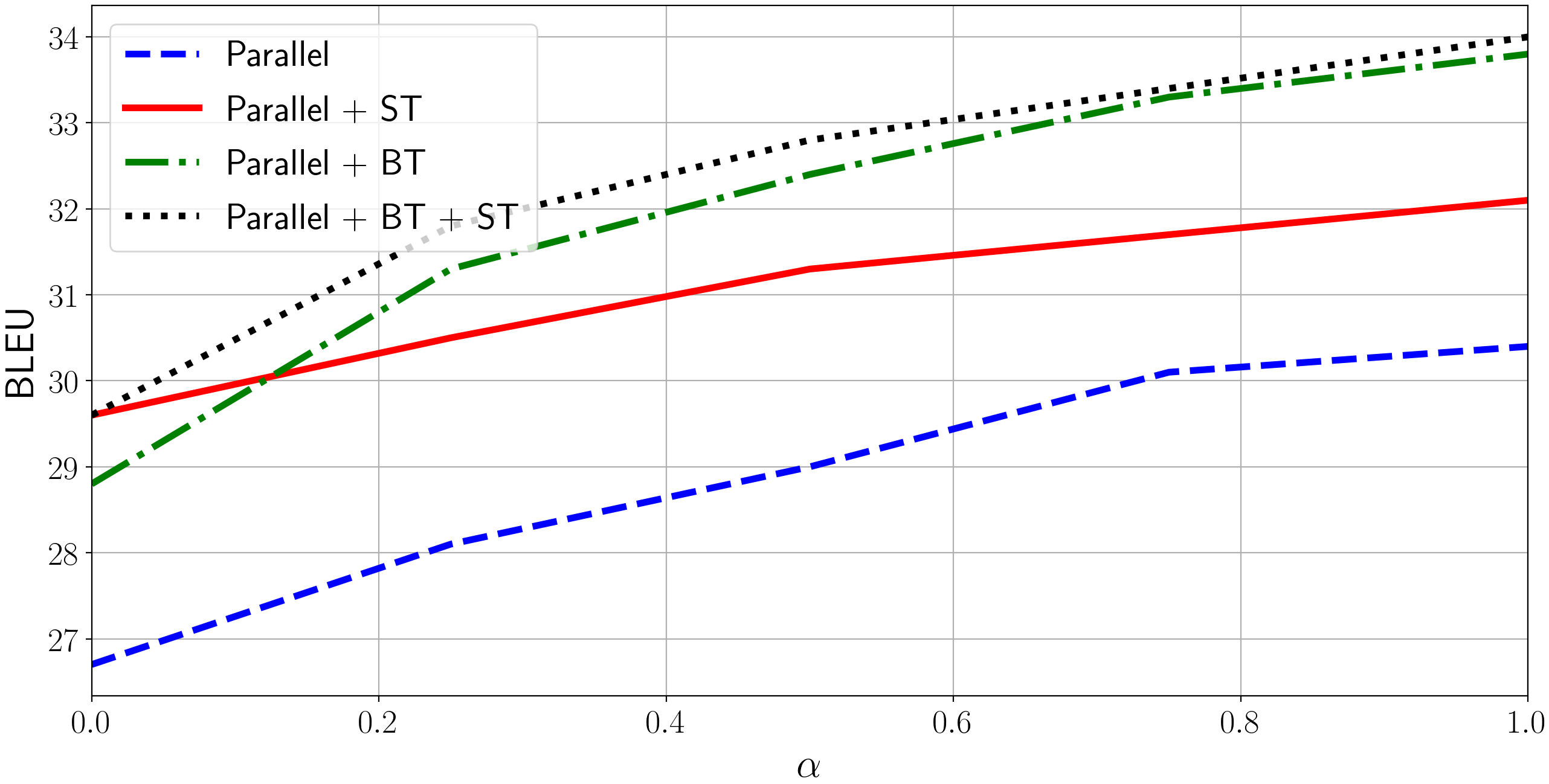}
\caption{\small BLEU score in Fr-En as a function of the amount of STDM. The target domain is fully out-of-domain when
$\alpha=0$, and fully in-domain when $\alpha=1$.}
\label{fig:alpha}
\vspace{-.2cm}
\end{figure}

\subsection{Controlled Setting} \label{sec:mt_controlled}
In the default setting, we have a parallel dataset with 20,000 parallel sentences.
10,000 are in-domain source originating data (EuroParl) and the remaining 10,000 are target originating data from a mix of domains, controlled by $\alpha \in [0, 1]$: $\alpha$ EuroParl + $(1-\alpha)$ OpenSubtitles.
The source side monolingual dataset has 100,000 French sentences from EuroParl. 
The target side monolingual dataset has 100,000 English sentences from: $\alpha$ EuroParl + $(1-\alpha)$ OpenSubtitles.
Finally, the test set consists of novel French sentences from EuroParl which we translate in English.


\paragraph{Varying amount of STDM.}
In Fig.~\ref{fig:alpha}, we benchmark our baseline approaches while varying $\alpha$ (see \textsection\ref{sec:contr}), which
controls the overlap between source and target domain. 

First, we observe improved BLEU~\cite{bleu} scores for all methods as we increase $\alpha$.
Second, there is a big gap between the baseline trained on parallel data only and methods which leverage monolingual data.
Third, combining ST and BT works better than each individual method, confirming that these approaches are complementary.
Finally, BT works better than ST but the gap reduces as the target domain becomes increasingly different from the source
domain (small values of $\alpha$). In the extreme case of STDM ($\alpha=0$), ST outperforms BT. In fact,
we observe that the gain of BT over the baseline decreases as $\alpha$ decreases, despite that the amount of monolingual data and parallel data remains constant across these experiments, thus showing that \emph{BT is less effective in the presence of STDM}.

\begin{figure}
\centering
\includegraphics[width=1\linewidth]{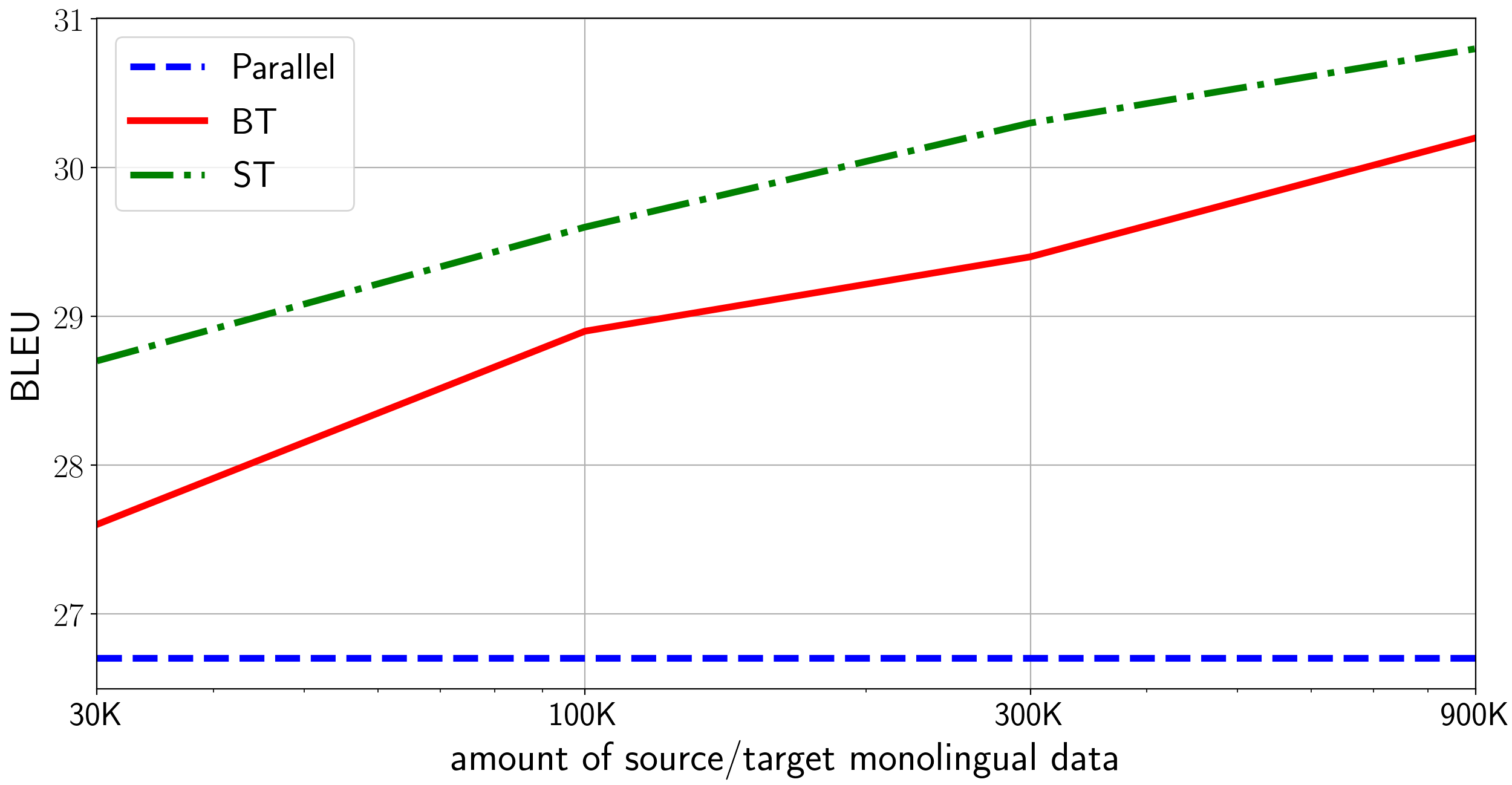}
\caption{\small BLEU as a function of the amount of monolingual data when $\alpha=0$.}
\label{fig:mono}
\vspace{-.2cm}
\end{figure}

\paragraph{Varying amount of monolingual data.}
We next explore how the quantity of monolingual data affects performance and if the relative gain 
of ST over BT when $\alpha=0$ disappears as we provide BT with more monolingual data. 
The experiment in Fig.~\ref{fig:mono}
shows that a) the gain in BLEU tapers off exponentially with the amount of data (notice the log-scale in the x-axis), 
b) for the same amount of monolingual data ST is always better than BT and by roughly
the same amount, and c) BT would require about 3 times more target monolingual data (which is out-of-domain) to yield 
the performance of ST. Therefore, \emph{increasing the amount of data can compensate for domain mismatch}.

\begin{figure}
\centering
\includegraphics[width=1\linewidth]{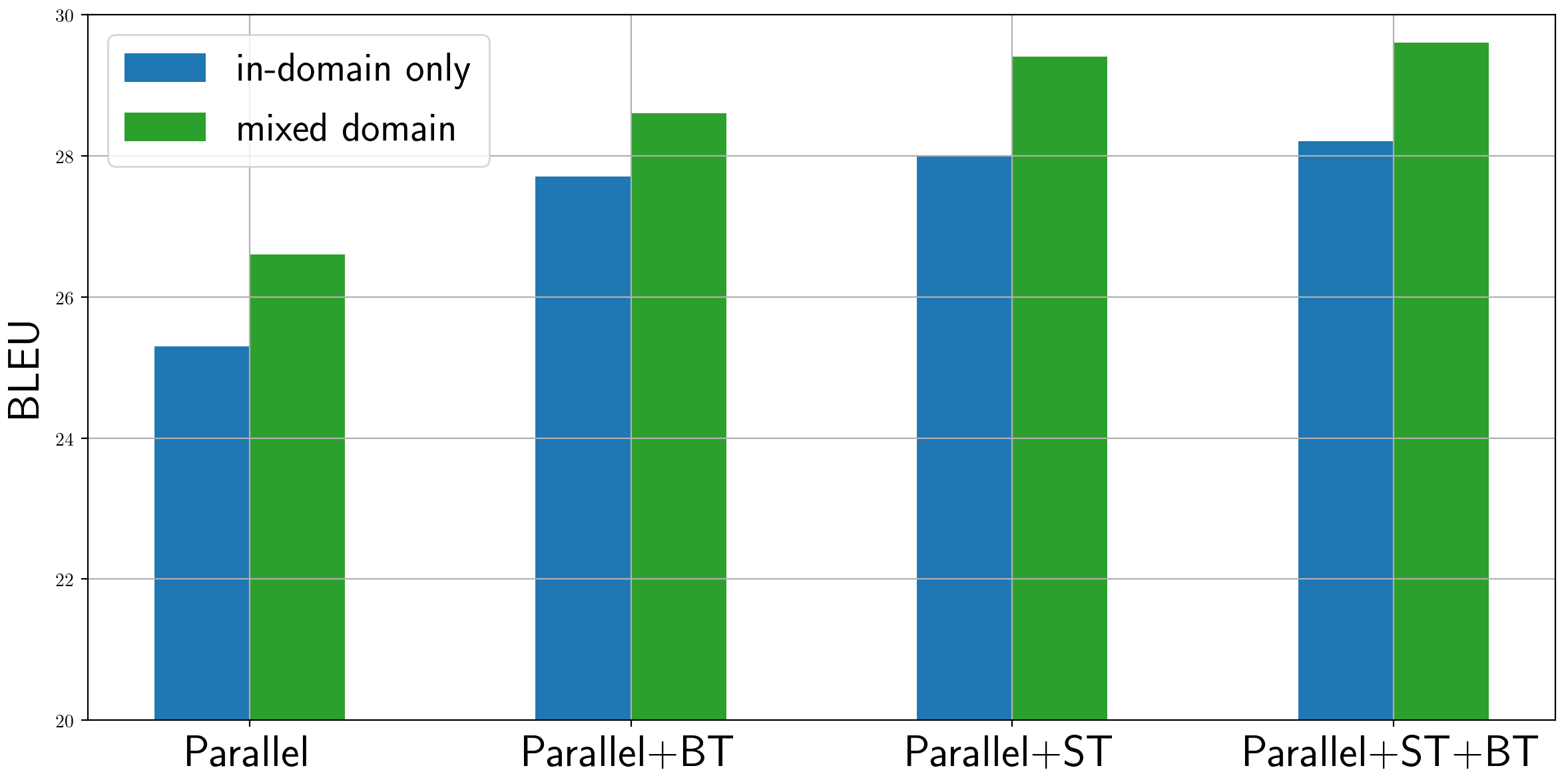}
\caption{\small BLEU when using only source originating in-domain data 
(blue bars) or also out-of-domain target originating data (green bars) for $\alpha=0$.}
\label{fig:indomainVSmixed}
\vspace{-.2cm}
\end{figure}

\paragraph{Varying amount of in-domain data.}
Now we explore whether, in the presence of extreme STDM ($\alpha=0$), it may be worth restricting the training data 
to only contain in-domain source originating sentences. In this case, the parallel set is reduced to 10,000 EuroParl sentences, the target side monolingual data is removed and back-translation is performed on the target side of the parallel dataset.
Fig.~\ref{fig:indomainVSmixed} demonstrates that in all cases
it is better to include the out-of-domain data originating on the target side (green bars).
Particularly in the low resource settings considered here, 
\emph{neural models benefit from all available examples even if these are out-of-domain}.

\begin{figure}
\centering
\includegraphics[width=1\linewidth]{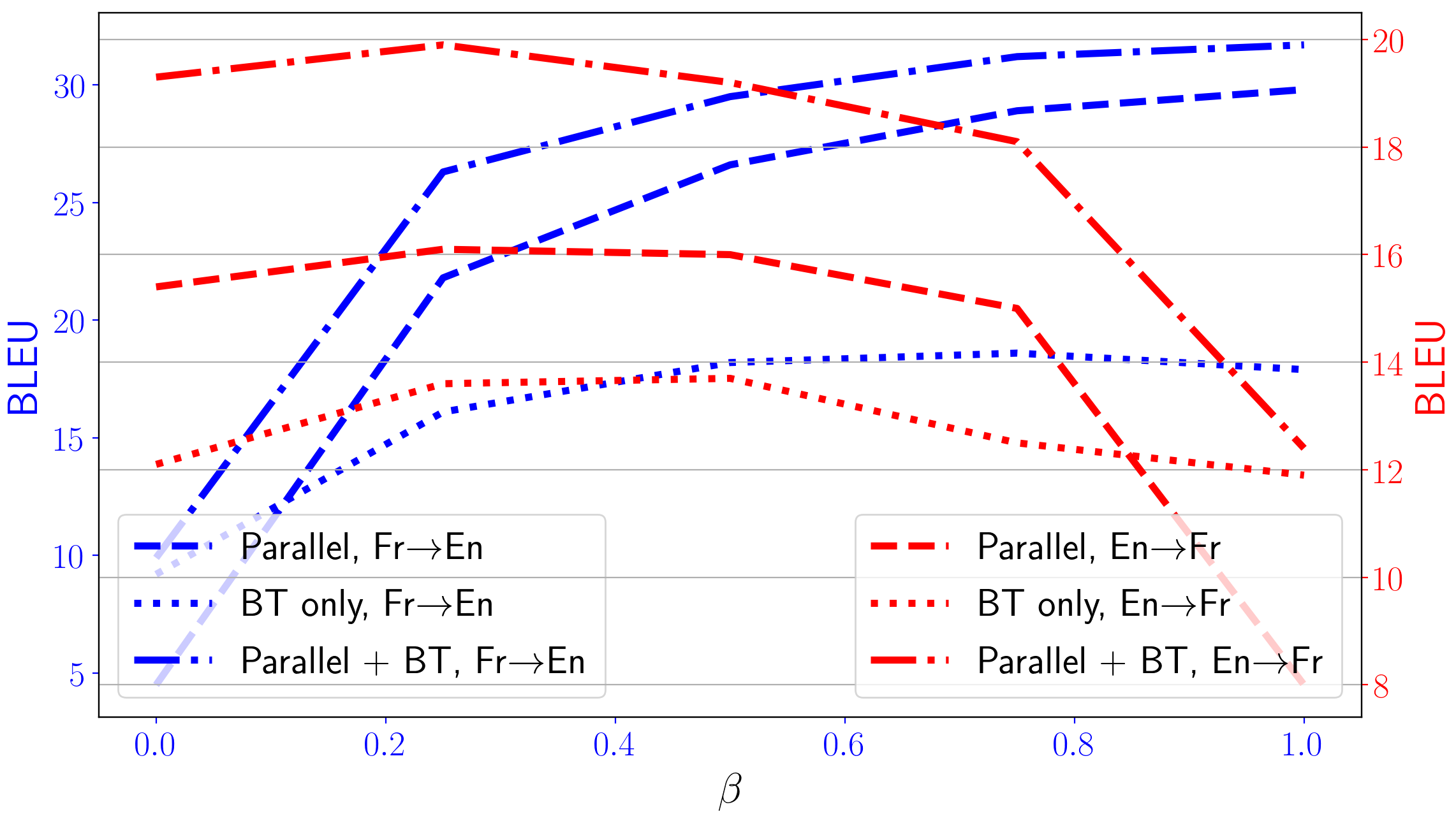}
\caption{\small BLEU score as a function of the proportion of parallel data originating in the source and target domain domain.
When $\beta=0$ all parallel data originates from OpenSubtitles, 
when $\beta=1$ all parallel data originates from EuroParl. Source and target monolingual corpora have 900,000 sentences 
from EuroParl and OpenSubtitles, respectively. 
The blue curves show BLEU in the forward direction (Fr-En translation of EuroParl data).
The red curves show BLEU in the reverse direction (En-Fr translation of OpenSubtitles sentences).
}
\label{fig:beta}
\vspace{-.2cm}
\end{figure}

Finally, we investigate how to construct a parallel dataset when STDM is significant ($\alpha=0$), i.e.
 the target domain is OpenSubtitles.
If we have a translation budget of 20,000 sentences, is it best to translate 20,000 sentences from EuroParl or to also include sentences from OpenSubtitles?
This is not obvious when training with BT, since the backward model may benefit from in-domain OpenSubtitles data.
In order to answer this question, we consider a {\em parallel} dataset with 20,000 sentences defined as: 
$\beta$ EuroParl + $(1-\beta)$ OpenSubtitles, with
$\beta \in [0, 1]$. When $\beta=0$, the parallel dataset is out-of-domain; when $\beta=1$ the parallel data is all in-domain.
The target side monolingual dataset is fixed and contains 900,000 sentences from OpenSubtitles.

Fig.~\ref{fig:beta} shows that taking \emph{all} sentences from EuroParl ($\beta=1$) is optimal 
when translating from French (EuroParl) to English (blue curves).
At high values of $\beta$, we observe a slight decrease in accuracy for models trained only on back-translated data (dotted line), confirming that BT loses its effectiveness when the reverse model is trained on out-of-domain data.
However, this is compensated by the gains brought by the additional in-domain parallel sentences (dashed line).
In the more natural setting in which the model is trained on both parallel and back-translated data (dash-dotted line), 
we see monotonic improvement in accuracy with $\beta$. A similar trend is observed in the other direction (English to French, red lines).
Therefore, if the goal is to maximize translation accuracy in {\em both} directions, an intermediate value of 
$\beta$ ($\approx 0.5$) is more desirable. 

\subsection{Low-Resource MT} \label{sec:lrmt}
We now test our approaches on two low-resource language pairs,
Nepali-English (Ne-En) and English-Myanmar (En-My). Nepali and Myanmar are spoken in regions with unique local context
that is very distinct from English-speaking regions, and thus these make good language pairs for studying the STDM setting in real life.

\paragraph{Data.} 
The Ne-En parallel dataset is composed of 40,000 sentences originating in Nepali and only 7,500 sentences originating in English.
There are 5,000 sentences in the validation and test sets all originating in Nepali.
We also have 1.8M monolingual sentences in Nepali and English, collected from public posts from a social media platform.
This dataset closely resembles our idealized 
setting of Fig.~\ref{fig:outline2}. The STDM score of this dataset is 0.64 (see Tab.~\ref{tab:stdm_score_real_ds})
and is analogous to our synthetic setting (\textsection\ref{sec:mt_controlled}) where $\alpha$ is low but $\beta$ is large.

The En-My parallel data is taken from the Asian Language Treebank (ALT) 
corpus \citep{alt,ding2018nova,ding2019towards} with 18,088 training sentences all originating from English news.
The validation and test sets have 1,000 sentences each, all originating from English. 
Following~\citet{chen19wat}, we use 5M English sentences from NewsCrawl as source side monolingual data and 100K Myanmar sentences
from Common Crawl as target side monolingual data.
We cannot compute an STDM score (\textsection\ref{sec:metric}) since we have no parallel data originating in Myanmar. Comparing
to our controlled setting this dataset would have $\beta$ equal to 1 and presumably a small value of $\alpha$, an ideal setting for ST.

\paragraph{Models.}
On both datasets, the parallel data baseline is a 5-layer transformer with 8 attention heads, 512 embedding dimensions and 2048 inner-layer dimensions, which consists of 42M parameters.
When training with BT and ST, we use a 6-layer transformer with 8 attention heads, 1024 embedding dimensions, 2048 inner-layer dimensions, resulting in 186M parameters.

\begin{table}[t]
\centering
\begin{tabular}{l||c|c}
\hline
Model &  Ne $\rightarrow$ En & En $\rightarrow$ My\\
\hline
baseline & 20.4 & 28.1 \\
BT & 22.3 & 30.0 \\
ST  & 22.1 & 31.9 \\
ST + BT & 22.9 & 32.4 \\
\end{tabular}
\caption{\small BLEU scores for the  Nepali to English and English to Myanmar translation task.}
\label{tab:my_ne_result}
\end{table}

\paragraph{Results.}
In Table \ref{tab:my_ne_result}, we observe that on the Ne-En task augmenting the parallel dataset 
with either forward- or back-translated monolingual data achieves almost 2 BLEU points improvement over the supervised baseline. 
On the En-My task BT slightly outperforms the baseline, while ST 
improves by $+2.5$ BLEU, since source side monolingual data is in-domain with the test set, while target side monolingual data is scarce and
out-of-domain.  On both tasks, we observe that combining ST and BT outperforms each individual method.

\section{Practical Tips}
Given these considerations and findings, how can we best set up a machine traslation system on a distant and possibly low-resource language 
pair? Our first recommendation is to be aware of possible STDM, and {\it(i)} 
check whether origin language information is available. If this is available, then it may be possible
to {\it(ii)} qualitatively look at the data to assess the extent of STDM, 
and quantitatively measure STDM as described in \textsection\ref{sec:metric}. 
Next, {\it(iii)} be aware that when STDM is severe, BT performance suffers (Fig.~\ref{fig:alpha}). However, 
{\it(iv)} we may be able to combat this 
by increasing the amount of target side (out-of-domain) monolingual data (Fig.~\ref{fig:mono}) and {\it(v)} 
by combining BT with ST (Fig.~\ref{fig:alpha}). 

Of course, the relative ratio of monolingual data in the source and target side 
and the actual degradation brought by STDM depend on the particular language pair. The more distant are the two languages, the more
difficult the learning task and the more data is needed to learn it. And finally, the less parallel data there is, the more monolingual
data will be needed to compensate. Therefore, there is an overall intricate dependency between all these factors, 
which we currently do not have neither theoretical nor practical tools to analyze and which certainly merits future investigation.

\section{Final Remarks} \label{sec:conclusions}
In this work we introduced the problem of source-target domain mismatch in machine translation.
We have formally defined STDM (\textsection\ref{sec:stdm}) 
and proposed a practical method to measure it (\textsection\ref{sec:metric}). While the commonly used WMT
datasets exhibit mild STDM, we find that less curated datasets in more distant and often lower resource language pairs (\textsection\ref{sec:stdm_score_ds}) exhibit much stronger STDM. We then investigated the effects of STDM on commonly used algorithms for training machine
translation systems and conclude that popular methods like BT are indeed affected.
Looking forward, we are interested in investigating better approaches to analyze and cope with STDM, 
to extend this study to the more realistic multilingual setting with multiple domains,
and to build public benchmarks that exhibit this natural phenomenon.

\section{Acknowledgments}
The authors would like to thank Marco Baroni, Silvia Bernardini, Randy Scansani, Alberto Barr\'on-Cede\~no, Adriano Ferraresi, and
Adina Williams for pointing to relevant references in the socio-linguistic literature and for general suggestions.
They also wish to thank Sergey Edunov for various tips on training MT systems at scale.

\bibliography{tacl2018}
\bibliographystyle{acl_natbib}

\end{document}